\documentclass{article}

\PassOptionsToPackage{numbers, compress}{natbib}

\usepackage[preprint]{neurips_2024}

\usepackage[utf8]{inputenc}
\usepackage[T1]{fontenc}
\usepackage{hyperref}
\usepackage{url}
\usepackage{booktabs}
\usepackage{amsfonts}
\usepackage{nicefrac}
\usepackage{microtype}
\usepackage{xcolor}
\usepackage{graphicx}
\usepackage{multirow}
\usepackage{placeins}

\usepackage{enumitem}
\usepackage{amsmath}

\title{Deep Temporal Deaggregation:\\Large-Scale Spatio-Temporal Generative Models}

\author{
	David Bergström \\
	Linköping University \\
	Linköping, Sweden \\
	\texttt{david.bergstrom@liu.se} \\
	\And
	Mattias Tiger \\
	Linköping University \\
	Linköping, Sweden \\
	\texttt{mattias.tiger@liu.se} \\
	\And
	Fredrik Heintz \\
	Linköping University \\
	Linköping, Sweden \\
	\texttt{fredrik.heintz@liu.se} \\
}

\begin{document}

\maketitle

\begin{abstract}
	Many of today's data is time-series data originating from various sources, such as sensors, transaction systems, or production systems.
	Major challenges with such data include privacy and business sensitivity. Generative time-series models have the potential to overcome these problems, allowing representative synthetic data, such as people's movement in cities, to be shared openly and be used to the benefit of society at large.
	However, contemporary approaches are limited to prohibitively short sequences and small scales. Aside from major memory limitations, the models generate less accurate and less representative samples the longer the sequences are. This issue is further exacerbated by the lack of a comprehensive and accessible benchmark.
	Furthermore, a common need in practical applications is what-if analysis and dynamic adaptation to data distribution changes, for usage in decision making and to manage a changing world: What if this road is temporarily blocked or another road is added?
	The focus of this paper is on mobility data, such as people's movement in cities, requiring all these issues to be addressed. 
	To this end, we propose a transformer-based diffusion model, TDDPM, for time-series which outperforms and scales substantially better than state-of-the-art. This is evaluated in a new comprehensive benchmark across several sequence lengths, standard datasets, and evaluation measures.
	We also demonstrate how the model can be conditioned on a prior over spatial occupancy frequency information, allowing the model to generate mobility data for previously unseen environments and for hypothetical scenarios where the underlying road network and its usage changes. This is evaluated by training on mobility data from part of a city. Then, using only aggregate spatial information as prior, we demonstrate out-of-distribution generalization to the unobserved remainder of the city.
\end{abstract}

\section{Introduction}

Time-series data of human mobility is important because it enables pandemic forecasting and management \cite{ilin2021public}, smart city development \cite{wang2022deep}, urban governance \cite{xiong2024trajsgan}, human rights violation detection \cite{tai2022mobile} and monitoring of global migration induced by war and climate change \cite{niva2023world,alessandrini2020estimating}. Today there are two major challenges limiting the applicability of time-series data to these ends. The first is a \emph{shortage of publicly available data}~\cite{ansari2024chronos}. Data can only be collected and shared in limited capacity due to privacy concerns, business concerns and national security, creating a silo effect. There is also technical impracticality in collecting and sharing the massive volumes produced by streaming data sources \cite{della2009s}.

Secondly, \emph{predictions} about unobserved parts in space, about the future or even possible futures - given that different actions are taken - is often a necessary complement to the readily collectable data. One such case is generating high-fidelity realistic spatio-temporal mobility data, such as individual pedestrians navigating a city or a building. Long-term prediction is essential for logistics planning, macroscopic network planning, infrastructure development, as well as useful for advanced traffic management systems (ATMS) \cite{lana2018road}. Open problems within the road traffic domain \cite{lana2018road}, and using human mobility data at large, is (1) high quality large-scale long-term predictions, (2) adaptation to sudden environmental changes and (3) facilitation of what-if analysis for environmental changes.

A promising solution to both of these challenges is to use time-series generative models to accurately capture and sample unknown time-series distributions. These generative models can be adapted to be trained and sampled to generate samples that are private~\cite{yoon2018pategan,wang2023pategail},
resulting in synthetic non-private datasets that can be made publicly available. The resulting models can also be adapted for tasks such as imputation~\cite{alcaraz2023diffusionbased}, forecasting~\cite{alcaraz2023diffusionbased}, and as basis for time-series foundation models~\cite{ansari2024chronos}.

However, current approaches have crucial limitations limiting their real-world applicability: they can only generating very short sequence length and they struggle to model complex distributions with sparse support.
Previous work on unconditional generation of time-series data has focused on sequence lengths of 24~\cite{yoon2019time}, 100~\cite{seyfi2022generating} and 256~\cite{yuan2024diffusionts}, and struggle to model even the smaller of spatio-temporal datasets used in this work. Furthermore, current methods have severe degrading performance with increased prediction length, which is a problem in the road traffic domain \cite{lana2018road}. 

Current approaches also struggle with environmental and context changes. The environment (and how people behave within it) often undergo rapid changes. This necessitates \emph{data ageing mechanisms that allow models to adapt their prediction to changing circumstances} \cite{lana2018road}, such as for example road construction, traffic accidents, traffic-light malfunction or other traffic flow and road topology changes. For analysis and what-if scenario modeling, it is further important that the distributions of synthetic data is proportional to the probability distribution of the real data: Certain locations and motion patterns are more frequently occurring than others. This is important for policy making, planning and decisions to be resting on correct risk assessments using for example Bayesian inference.

In this work, \textit{we introduce a new method for generating long and realistic sequences of spatio-temporal data, Temporal Denoising Diffusion Probabilistic Model (TDDPM)}, capable of out-of-distribution generalization via deaggregation from spatial statistics to temporal time-series samples.
More specifically, we first adapt the denoising diffusion model to generate time-series data and show that it scales to significantly longer sequences than what is possible with previous models.
Secondly, we ask: \emph{Can a deep generative model (in our case a diffusion model) be made to stay true to a marginal probability distribution, while generating samples from the full distribution?}
We demonstrate how this can be achieved via conditioning the model on a non-temporal marginal distributions of the trajectory data, making deaggregation to individual trajectories possible. A consequence of this is that the model can generalize to environment changes and even to new environments without retraining, requiring only a statistic (i.e., an aggregate) of the new environment as input.

\begin{figure}
    \centering
    \includegraphics[width=\textwidth]{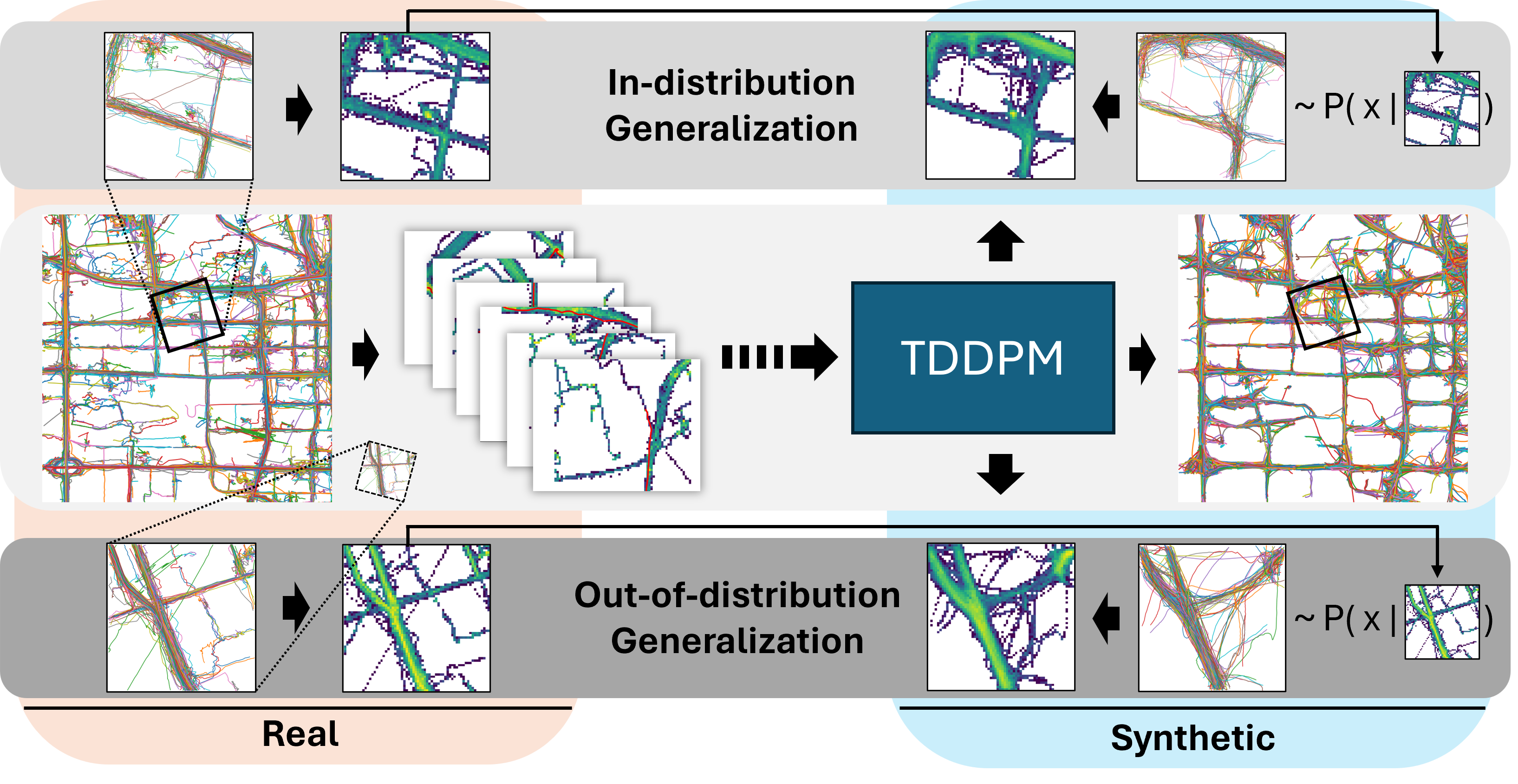}
    \caption{TDDPM is trained on real 2D trajectories (left) to generate synthetic trajectories (right), conditioned on how likely it should be for the population of synthetic trajectories to occupy space on the 2D plane. The latter is represented as a discrete distribution over occupancy frequency, i.e., the marginal distribution over the trajectory probability distribution if integrating out time. This yields both high-fidelity in-distribution generalization (top) and out-of-distribution generalization (bottom), the latter when conditioning on a marginal distribution not part of the training data (dashed rectangle). }
    \label{fig:overview}
\end{figure}

Finally, \textit{we introduce an hierarchical approach where a large geographical region is sampled, by sampling several smaller sub-regions using a single conditional denoising diffusion model}.
This is achieved by conditioning the denoising diffusion model on local information. The local information is statistics extracting about a local area which is translation and rotation invariant. This allow our approach to scale to city-size problems, without compromising fidelity, support nor proportionality of the distribution.
Figure \ref{fig:overview} illustrates the full capabilities of TDDPM in a setting of mobility data.

Our contributions are (1) a hierarchical method for generating large-scale high-fidelity spatio-temporal data using conditional denoising diffusion, and (2) demonstrating that the new model remains competitive with other unconditional time-series generative model.

\section{Problem Definition}

The task of learning an unconditional generative model is defined as learning a mapping $f$ from samples drawn from a known distribution $\mathcal{D}_\mathrm{known}$, e.g. standard normal distribution, to samples from an unknown target distribution $\mathcal{D}_\mathrm{unknown}$. 
The mapping is learned without direct access to the unknown distribution, and is instead limited to a set of samples $X_\mathrm{train} = \{x_1, \dots, x_N\}$, where $x_i \sim \mathcal{D}_\mathrm{unknown}$.
Once the mapping has been learned, synthetic data can be generating by first sampling the known distribution, then passing the individual samples through the mapping function $X_\mathrm{synthetic} = \{f(y_i)\}_{i=0}^M$, where $y_i \sim \mathcal{D}_\mathrm{known}$.
The goal of this mapping is for the synthetic samples to be \textit{similar} to samples from the known distribution.
The notion of similarity is often broken into several desirable properties for the mapping function and the resulting synthetic data points, in this work we consider the following properties:
\begin{enumerate}[label=\Roman*.]
	\item \textit{Fidelity:}~\cite{alaa2022faithful} The individual synthetic samples should have similar characteristics to, or be indistinguishable from, samples from the original distribution. 
	\item \textit{Diversity:}~\cite{alaa2022faithful} It should be possible for synthetic data to be drawn from any part of the unknown distribution's support.
	\item \textit{Proportionality:} \cite{wu2021bridging} The probability of a sample occurring in the synthetic distribution should be proportional to the probability of a sample occurring in the unknown distribution.
	\item \textit{Usefulness:}~\cite{esteban2017real} The synthetic data should capture aspects of the unknown distribution that is useful for downstream tasks.
	\item \textit{Generalization:}~\cite{alaa2022faithful} Synthetic samples should not be copies of the training data.
\end{enumerate}
Note that these attributes are not necessarily independent, e.g. creating a proportional synthetic dataset (III) can be difficult without the ability to generate diverse samples (II).

In this work, we separate the generative task into two steps by extracting local information $l$ from the training data.
We can then condition the generation on this local information.
This allows for more accurate modelling on challenging distributions, as well as enabling generalization in some cases where $l$ can include sufficient information. 
More precisely, instead of learning to generate samples from $p(x)$ directly, we learn to generate samples from the joint distribution $p(x, l)$ via the chain rule: 
\begin{equation}
	p(x, l) = p(x | l) \, p(l),
\end{equation}
where $p(x | l)$ is a conditional deep generative model and $p(l)$ can be modeled using a explicit and more interpretable distribution.

Creating synthetic samples from $p(x)$ is then done by first sampling $p(l)$ and then use these samples to sample the conditional distribution $p(x | l)$.
In practice, we learn a conditional mapping function $g$ which maps samples from a known distribution and local information $l$ to synthetic samples, which should be \textit{similar} to samples from the unknown distribution $\mathcal{D}_\mathrm{unknown}$.
More precisely, a dataset for a collection of local information $L = \{l_i\}_{i=0}^M$ is generated $X_\mathrm{synthetic} = \big\{g(y_i, l_i)\big\}_{i=0}^M$, where $y_i \sim \mathcal{D}_\mathrm{known}$.

For certain problems, if the choice of $l$ is sufficiently informative, we can learn to generate data for a new distribution $p^*(x)$ without having to retrain $p(x | l)$.
This is achieved by estimating a new distribution $p^*(x, l)$, s.t.
\begin{equation}
	p^*(x, l) \approx p(x | l) \, p^*(l).
\end{equation}
This is not only a computationally efficient way to estimate $p^*(x)$, but can also be useful in cases where we only have access to $l$, e.g. to model scenarios or account for data distribution shifts. 

\section{Related Work}

Previous work on unconditional generation of time-series data has focused on variations of the generative adversarial networks (GAN) architecture~\cite{esteban2017real,yoon2019time,jeon2022gt} and, more recently diffusion models~\cite{yuan2024diffusionts}.
There has also been an interest in using time-series generation for imputation and forecasting~\cite{tashiro2021csdi,shen2023non,alcaraz2023diffusionbased,dai2023timeddpm,feng2024latent}.
Transformer-based time-series foundation models has been proposed as a general purpose forecasting tool~\cite{ansari2024chronos}, but has not been evaluated on the unconditional generation task.

TimeGAN~\cite{yoon2019time} consists of a generative adversarial network (GAN) operating inside the latent space of an autoencoder.
To further improve performance, they add an additional network with the task of predicting one time step ahead.
The encoder, decoder, supervisor, generator and discriminator are all implemented using autoregressive models and in practice they use gated recurrent units (GRUs). 
The TimeVAE~\cite{desai2021timevae} architecture is a variant of the popular variational autoencoder architecture.
The autoencoder is trained with an additional loss component to have the latent space conform to a known statistical distribution, in this instance a multivariate normal distribution.
The autoencoder is trained to both minimize the reconstruction loss, as well as minimizing the divergence between the embedded data and the prior set for the latent space.
COSCI-GAN~\cite{seyfi2022generating} proposes to use a separate generative adversarial network for each channel of the data. The individual GANs all share single source of noise as input to the generator and, additionally they a central discriminator that is given the stacked output from the all the generators as input.

Denoising diffusion probabilistic models have also been proposed as an alternative to both GAN and VAE architectures.
DiffusionTS~\cite{yuan2024diffusionts} adapts this architecture to generate time-series data by implementing the denoising step with a multilayer neural network each consisting of a transformer block, a fully connected neural network as well as time-series specific layers with the aim of improved interpretability.

The transformer architecture~\cite{vaswani2017attention} proposed attention as an alternative to recurrence in neural networks and the architecture itself became a common architecture.
The original architecture consists of a encode and a decoder, but for certain tasks one of the two is often sufficient.
The vision transformer (ViT)~\cite{dosovitskiy2020image} relies on the transformer encoder and splits images into several equally sized patches which are projected before given as input tokens to the encoder.

There is also a community dedicated to the specific problem of learning generative trajectory models.
Most models are variations on the GAN architecture~\cite{feng2020learning,jiang2023continuous,han2024enhanced,xiong2024trajsgan}.
The COLA architecture~\cite{wang2024cola} is based on the transformer architecture, with support for re-tuning the network to generate trajectories for different cities.
Our model can generalize to new parts of the same city, something that is possible with the COLA architecture, but our proposed model does not require any re-tuning.
TS-TrajGen~\cite{jiang2023continuous} generates continuous trajectories by applying A* to an explicit traversable representation of the road network provided as input.
Our proposed model uses a discretized marginal distribution of the movement to guide the generation process, this allows us to generate trajectories in areas where map data is not available, e.g. in indoor environment or forest areas where roads are less concretely defined.
The discretized marginal distributions also contain information about which roads are more commonly travelled and which have less traffic.
This allows for more representative synthetic trajectory datasets.

In this work, we focus on using denoising diffusion to generate time-series data.
This is similar to the work of DiffusionTS~\cite{yuan2024diffusionts}, but we introduce less inductive bias by using a simpler architecture without any time-series specific layers, and instead we rely on the time-embedding of the original transformer architecture~\cite{vaswani2017attention}.
For the conditional mode, we condition our model on a prior which in practice results in heatmaps being given as an additional input.
Similar to ViT, we split our heatmaps into 8x8 equally sized patches and use linear projection to convert them into tokens.

\section{Temporal Denoising Diffusion Probabilistic Model (TDDPM)}

\begin{figure}
	\centering
	   \includegraphics[width=0.6\textwidth]{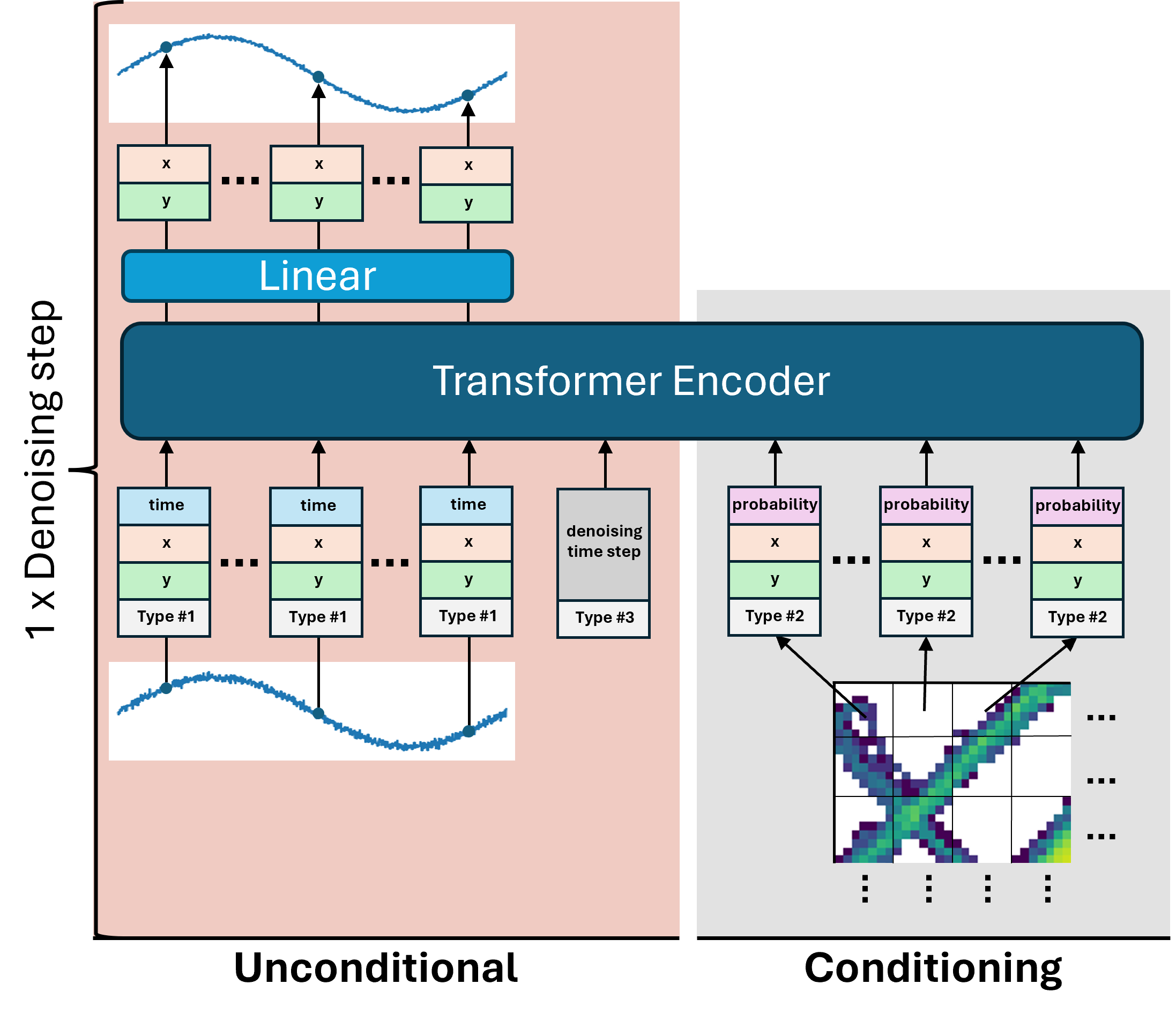}
	\caption{Overview of the architecture. In the unconditional part, each time point of the noisy trajectory is converted into a separate token with positional embedding~\cite{vaswani2017attention} used to embed its values and the time point, as well as a learned vector representing its type. The denoising step token encodes the denoising step, the step is encoding using positional encoding and then concatenated with a type vector. The transformer can also optionally take a marginal distribution to guide the denoising process to generate samples with particular properties, improving in-distribution performance as well as enabling generalization to previously unobserved areas. The marginal distribution is split into tokens, taking inspiration from ViT~\cite{dosovitskiy2020image}, concatenated with a learned type vector and the corresponding position using positional embedding.}
	\label{fig:architecture}
\end{figure}

The architecture is based on the denoising diffusion architecture~\cite{ho2020denoising}, using a transformer encoder~\cite{vaswani2017attention} for denoising.
An overview of the approach is shown in Figure \ref{fig:architecture}.
The model can run in two modes: (1) unconditional and (2) conditional mode. In (2) we condition the generation on aggregated information.
In both modes, we use the denoising diffusion architecture to generate entire sequences of time-series data.
At each step of the denoising process, the transformer encoder is fed a noisy time-series sequence as well as the current step and is tasked to predict the noise added at the given time step.
In the conditional mode, the transformer encoder is also fed additional tokens containing the aggregated information (\ref{appendix:encoder-input}).

\subsection{Denoising diffusion}

Next, we provide a short overview of the denoising diffusion architecture.
For more details and intuition, we refer to the original paper~\cite{ho2020denoising}.
Denoising diffusion models generate data by learning to invert a known noise-adding process, called the forward process:
\begin{equation}
	q_\theta(x_t \vert x_{t-1}) := \mathcal{N}(x_t ; \sqrt{1 - \beta_t}x_{t-1}, \beta_t I)
\end{equation}
For the reverse process, they parameterize reverse process like this:
\begin{equation}
	p_\theta(x_{t-1} \vert x_t) = \mathcal{N}(x_{t-1} ; \mu_\theta(x_t, t), \Sigma_\theta(x_t, t)),
\end{equation}
where $\Sigma_\theta(x_t, t) = \sigma_t^2 I$ and 
\begin{equation}
	\mu_\theta(x_t, t) = \frac{1}{\sqrt{\alpha_t}} \left( x_t - \frac{\beta_t}{\sqrt{1-\bar{\alpha}_t}}\epsilon_\theta(x_t, t) \right).
\end{equation}
where $\epsilon_\theta$ is approximated using a neural network.
In the original setting, they propose a simplified training objective for the reverse process:
\begin{equation}
	L(\theta) := \mathbb{E}_{t, x_0, \epsilon} \left[ \left\Vert \epsilon - \epsilon_\theta \left( \sqrt{\bar{\alpha}}x_0 + \sqrt{1-\bar{\alpha}_t} \epsilon, t \right) \right\Vert^2 \right],
\end{equation}
where $t$ is sampled uniformly between $1$ and $T$ during training.

\subsection{Architecture}

In this work we extend $\epsilon_\theta$ to take an additional argument $l$ containing local information  to the area we are generating. 
The sampling is done in two steps: (1) Calculate some local statistic $l$ about the local environment, in this case a heatmap of the data in a local area. (2) Condition the time-series generation on said statistics to generate samples local to that area.
By repeating the two steps for different local areas, we can aggregate several local samples to form a representative sample for the full geographical distribution.
The training data consists of a set of $N$ trajectories, where each trajectory $t_i = \{ p_1, p_2, \dots, p_{L_i} \}$, where each observation $P^D$.

We set $l$ to be a local heatmap of the training data trajectories for a given region $r = (x, y, \theta)$ where $\theta$ is rotation in [$0$, 2$\pi$).
A few example heatmaps are shown in figure \ref{fig:overview}.
During training, we sample these regions uniformly (including rotation) and extract pairs of heatmap and sub-trajectories in the region.
The sub-trajectories are transformed by subtracting the regions origin point and by rotating it appropriately.
If the sub-trajectory is longer than the target sequence length, a random sub-sequence is chosen.
The heatmaps contain information about relative frequency, i.e. where which areas in the regions are more frequently visited and which areas are never visited.

When sampling, we condition the generation on heatmaps for the region we want to generate for.
This allows for both interpolation, i.e. generating for regions that are not part of the training set, but that are from the same physical space and time, as well as extrapolation, generating sequences that are outside of the distribution.

Since the transformer architecture is permutation invariant, positional encoding~\cite{vaswani2017attention} is used to introduce order when necessary.
In this work we use the variant of the original positional encoding introduced in the original work on denoising diffusion probabilistic models~\cite{ho2020denoising} shown in Equation \ref{eq:positional-encoding}.

\section{Evaluation}

In this section we first study the performance in unconditional mode.
We evaluate our proposed method on the unconditional generation task by comparing it to several state of the art methods across multiple standard datasets and sequence lengths. The four unconditional methods are representative of GAN, VAE and Diffusion approaches: TimeGAN \cite{yoon2019time}, TimeVAE \cite{desai2021timevae}, COSCI-GAN \cite{seyfi2022generating} and DiffusionTS \cite{yuan2024diffusionts}. 
Note that a separate model is trained for each combination of sequence length and dataset. Then we evaluate TDDPM in conditional mode, which the other methods do not support. We validate out-of-distribution generalization performance of TDDPM in the large-scale spatio-temporal task of time series generative modelling of mobility data. Finally, we demonstrate what-if scenario modelling and showcase out-of-distribution generalization when conditioned to synthetic local information. 

The efficiency of training, re-training and sampling is relevant to how and when different time series generative models can be used in different applications, and for sustainable use of ML. Training TDDPM is highly efficient (Table \ref{table:time-inefficiency}). It is up to 1,000x faster than GAN-based SOTA, 50x faster than VAE-based SOTA and 6x faster than DiffusionTS. Interestingly, DiffusionTS is increasingly slower w.r.t. TDDPM as the sequence length increases. As expected \cite{xiao2021tackling}., the diffusion-based methods take much longer to sample than the GAN and VAE based method  In between the diffusion-based methods, TDDPM is 250x faster at sampling than DiffusionTS. TDDPM takes approximately 1.15 minutes to draw $10000$ samples.

\begin{table*}[t]
  \label{table:time-inefficiency}
  \caption{Time inefficiency, as a factor of TDDPM training time and sampling time, respectively. Training time are shown in intervals (lowest to highest across datasets). 10,000 samples are drawn. }
    \centering
    \small
    \begin{tabular}{l ccccc}
        \addlinespace[0.5em]
        \toprule
        & TimeGAN & TimeVAE & COSCI-GAN & DiffusionTS & TDDPM \\
        \midrule
        \addlinespace[0.5em]
        Training & 200 - 900 & 3 - 53  & 25 - 3510 & 2 - 6 & 1\\
        Sampling & 0.1 & 0.01 & 0.002 & 250 & 1\\
    \bottomrule
    \end{tabular}
\end{table*}

\subsection{Datasets}
We use 6 real-world datasets, where Stock \cite{yoon2019time}, Energy \cite{yoon2019time}, Solar \cite{jeha2021psa} and Electricity \cite{jeha2021psa} are established in the generative time series field. The last two, ATC \cite{brvsvcic2013person} and GeoLife \cite{zheng2011geolife} are large-scale mobility data sets from the robotics field.
The datasets used are summarized in Table \ref{table:datasets}.
The ATC shopping center dataset consist of pedestrians being tracked indoors between 2012 and 2013. We use the first day (2012/10/24) of the dataset, and drop all features except position (x,y) scaled to meters. Time is also dropped in favor for index since time between observations is roughly constant. GeoLife is a GPS trajectory dataset collected by Microsoft Research Asia, consists of 178 users and was collected between 2007 and 2011. The majority of the data is centered on Beijing, China. We use two subsets of the full GeoLife dataset: Geolife (Small) at 10 $\text{km}^2$ and Geolife (Large) at 161 $\text{km}^2$.

\subsection{Metrics}
The synthetic datasets from the trained models are compared to the original data using the following:

\textbf{TimeFID:} We adapt the widely used Fréchet Inception Distance (FID)~\cite{heusel2017gans}, a measure for evaluating image generative models, to time-series generation by replacing the pre-trained inception network with a domain-specific time-series embedding network~\cite{franceschi2019unsupervised}.
 This is similar to Context-FID~\cite{jeha2021psa}, however Context-FID is calculated for certain context, i.e. a window of time, while TimeFID is calculated for window.
 Prevalent numerical issues has also been addressed~\cite{higham1988computing}.

\textbf{TSTR:} Train on synthetic, test on real. We follow the evaluation done by TimeGAN~\cite{yoon2019time} and use the synthetic data to train a GRU-based RNN on task of one step prediction using the synthetic data. The resulting model is then evaluated on the training data and the mean absolute value is reported. This evaluates the usefulness, fidelity and diversity of the synthetic data. A lower score is better.

\textbf{t-SNE:} We refine the process of visual evaluation by using an embedding network to first embed the real and synthetic time-sequences into fixed-size vectors. The vectors are then passed to t-SNE which places the vectors on a 2D plane, similar vectors are placed close to each other and vectors that are different further away. The points are then colored, with different colors for the training and synthetic datasets. This allows us to visually evaluate the similarity between the synthetic and real data. More similar data are more similar in terms of fidelity and diversity, while having points that cover the entire range of real data points indicates support coverage. If all synthetic points share the same location as the training data, this suggests poor generalization.

\textbf{KL Divergence:} Evaluates support coverage as well as proportionality. KL(real$\,\vert\vert\,$synthetic) measures how well the synthetic distribution represents the real distribution, with $0$ being identical. KL(synthetic$\,\vert\vert\,$real) measures how well the synthetic distribution fits inside the real distribution, i.e. it is sufficient that the synthetic distribution matches a single mode of the real distribution in order to get a low divergence value. The synthetic distribution consequently do not need to have the same support, nor have any probability density outside of this mode. Symmetric KL weight these two together and Jensen–Shannon divergence is a more stable version of symmetric KL with regions without probability in either of the distributions.

\subsection{Unconditional Generation}

Evaluation using TimeFID (Table \ref{table:result-TimeFID}) show that TDDPM is on-par with top contenders, which vary between datasets), apart from for Solar and Electricity where TDDPM is better. TDDPM is also the only method which can scale to long sequences in all datasets. We also evaluated TSTR (\ref{table:result-TSTR}) for which most methods are on-par on most datasets. TSTR is consequently assumed to be saturated and need more complex tasks in future work.
In addition to the quantitative measures, we investigate the qualitative quality by plotting the training and synthetic datasets using t-SNE, see Figure \ref{fig:unconditional-tsne}. 
For GeoLife, we also calculate the spatial marginal distribution for the resulting synthetic trajectories, shown Figure \ref{fig:unconditional-geolife}.
These are then used to calculate the divergence between the synthetic and training data distributions, shown in Table \ref{table:KL} and Table \ref{table:divergence} (appendix). Low KL(real || synthetic) indicate that the synthetic distribution captures the real distribution well, which is distinctly seen for TDDPM. DiffusionTS, which is slightly worse, show better results for KL(synthetic || real), indicating that its synthetic distribution has captured some modes in the real data well, but not across the full support of the real distribution. The symmetric KL and JS divergence (\ref{table:divergence}) measures a compromise between these two, for which TDDPM and DiffusionTS tops and are on-par, with TDDPM being slightly stronger. Finally, in both tables we can see that conditional TDDPM on out-of-distribution generation outperforms all other methods (including unconditional TDDPM) with a wide margin.

\begin{table*}[t]
	\caption{Evaluation of unconditional generation using TimeFID. Lower is better. Several experiments ran out of 24GB of VRAM (\emph{oom}) or did not finish (\emph{dnf}) in under 100 hours.}
	\centering
	\small
	\begin{tabular}{lr ccccc}
		\addlinespace[0.5em]
		\toprule
		Dataset & Len. & TimeGAN & TimeVAE & COSCI-GAN & DiffusionTS & TDDPM \\
		\midrule

  \addlinespace[0.5em]
\multirow{7}{*}{Stock} & $24$ & $0.49 \pm 0.49$ & $0.15 \pm 0.15$ & $\mathbf{0.12 \pm 0.12}$ & $0.15 \pm 0.15$ & $0.13 \pm 0.13$ \\
                       & $32$ & $0.56 \pm 0.56$ & $0.13 \pm 0.13$ & $\mathbf{0.12 \pm 0.12}$ & $\mathbf{0.12 \pm 0.12}$ & $0.19 \pm 0.19$ \\
                       & $64$ & $0.25 \pm 0.25$ & $0.19 \pm 0.19$ & $0.21 \pm 0.21$ & $\mathbf{0.11 \pm 0.11}$ & $0.20 \pm 0.20$ \\
                       & $128$ & $2.40 \pm 2.40$ & $0.52 \pm 0.52$ & $0.22 \pm 0.22$ & $\mathbf{0.16 \pm 0.16}$ & $0.40 \pm 0.40$ \\
                       & $256$ & $1.87 \pm 1.87$ & $0.29 \pm 0.29$ & $0.14 \pm 0.14$ & $\mathbf{0.09 \pm 0.09}$ & $0.15 \pm 0.15$ \\
                       & $512$ & dnf & $1.37 \pm 1.37$ & $\mathbf{0.05 \pm 0.05}$ & oom & $0.28 \pm 0.28$ \\
                       & $1024$ & dnf & $130.71 \pm 130.71$ & $\mathbf{0.04 \pm 0.04}$ & oom & $0.30 \pm 0.30$ \\
\addlinespace[0.5em]
\multirow{7}{*}{Energy} & $24$ & $0.69 \pm 0.69$ & $0.43 \pm 0.43$ & $1.73 \pm 1.73$ & $\mathbf{0.27 \pm 0.27}$ & $0.40 \pm 0.40$ \\
                        & $32$ & $0.77 \pm 0.77$ & $0.44 \pm 0.44$ & $2.14 \pm 2.14$ & $\mathbf{0.30 \pm 0.30}$ & $0.45 \pm 0.45$ \\
                        & $64$ & $0.50 \pm 0.50$ & $0.35 \pm 0.35$ & $1.47 \pm 1.47$ & $\mathbf{0.22 \pm 0.22}$ & $0.37 \pm 0.37$ \\
                        & $128$ & $1.42 \pm 1.42$ & $0.52 \pm 0.52$ & $1.74 \pm 1.74$ & $\mathbf{0.19 \pm 0.19}$ & $0.51 \pm 0.51$ \\
                        & $256$ & $3.46 \pm 3.46$ & $0.50 \pm 0.50$ & $0.90 \pm 0.90$ & $\mathbf{0.21 \pm 0.21}$ & $0.52 \pm 0.52$ \\
                        & $512$ & dnf & $\mathbf{0.37 \pm 0.37}$ & $0.81 \pm 0.81$ & oom & $0.43 \pm 0.43$ \\
                        & $1024$ & dnf & oom & $0.56 \pm 0.56$ & oom & $\mathbf{0.37 \pm 0.37}$ \\
\addlinespace[0.5em]
\multirow{7}{*}{Solar} & $24$ & oom & $3.54 \pm 3.54$ & $2e8 \pm 2e8$ & $0.73 \pm 0.73$ & $\mathbf{0.66 \pm 0.66}$ \\
                       & $32$ & oom & $2.87 \pm 2.87$ & $2e7 \pm 2e7$ & $0.74 \pm 0.74$ & $\mathbf{0.48 \pm 0.48}$ \\
                       & $64$ & oom & $1.73 \pm 1.73$ & $4e6 \pm 4e6$ & $0.74 \pm 0.74$ & $\mathbf{0.53 \pm 0.53}$ \\
                       & $128$ & oom & $\mathbf{0.54 \pm 0.54}$ & $4e7 \pm 4e7$ & $0.88 \pm 0.88$ & $0.80 \pm 0.80$ \\
                       & $256$ & oom & oom & $4e7 \pm 4e7$ & $0.79 \pm 0.79$ & $\mathbf{0.38 \pm 0.38}$ \\
                       & $512$ & oom & oom & $4e7 \pm 4e7$ & oom & $\mathbf{0.52 \pm 0.52}$ \\
                       & $1024$ & oom & oom & dnf & oom & $\mathbf{1.24 \pm 1.24}$ \\
\addlinespace[0.5em]
\multirow{7}{*}{Electricity} & $24$ & oom & $5.27 \pm 5.27$ & dnf & $4.86 \pm 4.86$ & $\mathbf{2.29 \pm 2.29}$ \\
                             & $32$ & oom & $4.42 \pm 4.42$ & dnf & $4.49 \pm 4.49$ & $\mathbf{2.13 \pm 2.13}$ \\
                             & $64$ & oom & $4.65 \pm 4.65$ & dnf & $4.73 \pm 4.73$ & $\mathbf{1.35 \pm 1.35}$ \\
                             & $128$ & oom & oom & dnf & $4.73 \pm 4.73$ & $\mathbf{1.09 \pm 1.09}$ \\
                             & $256$ & oom & oom & dnf & $4.67 \pm 4.67$ & $\mathbf{1.52 \pm 1.52}$ \\
                             & $512$ & oom & oom & dnf & oom & $\mathbf{0.68 \pm 0.68}$ \\
                             & $1024$ & oom & oom & dnf & oom & $\mathbf{0.77 \pm 0.77}$ \\
\addlinespace[0.5em]
\multirow{7}{*}{ATC} & $24$ & dnf & $50.34 \pm 50.34$ & dnf & $\mathbf{0.13 \pm 0.13}$ & $0.14 \pm 0.14$ \\
                     & $32$ & dnf & $76.20 \pm 76.20$ & dnf & $\mathbf{0.14 \pm 0.14}$ & $0.19 \pm 0.19$ \\
                     & $64$ & dnf & $413.52 \pm 413.52$ & dnf & $\mathbf{0.11 \pm 0.11}$ & $0.18 \pm 0.18$ \\
                     & $128$ & dnf & oom & dnf & $\mathbf{0.14 \pm 0.14}$ & $0.16 \pm 0.16$ \\
                     & $256$ & dnf & oom & dnf & $\mathbf{0.22 \pm 0.22}$ & $0.24 \pm 0.24$ \\
                     & $512$ & dnf & oom & dnf & oom & $\mathbf{0.29 \pm 0.29}$ \\
                     & $1024$ & dnf & oom & dnf & oom & $\mathbf{0.42 \pm 0.42}$ \\
\addlinespace[0.5em]
 & $24$ & $1.05 \pm 1.05$ & $0.24 \pm 0.24$ & $0.29 \pm 0.29$ & $\mathbf{0.12 \pm 0.12}$ & $0.18 \pm 0.18$ \\
                                 & $32$ & $0.54 \pm 0.54$ & $0.41 \pm 0.41$ & $0.98 \pm 0.98$ & $\mathbf{0.16 \pm 0.16}$ & $0.19 \pm 0.19$ \\
                                 & $64$ & $0.36 \pm 0.36$ & $0.35 \pm 0.35$ & $0.27 \pm 0.27$ & $\mathbf{0.17 \pm 0.17}$ & $0.25 \pm 0.25$ \\
 Geolife                                & $128$ & $0.76 \pm 0.76$ & $0.21 \pm 0.21$ & $\mathbf{0.16 \pm 0.16}$ & $0.21 \pm 0.21$ & $0.21 \pm 0.21$ \\
 (Small)
                                 & $256$ & $0.96 \pm 0.96$ & $\mathbf{0.18 \pm 0.18}$ & $0.25 \pm 0.25$ & $0.28 \pm 0.28$ & $0.37 \pm 0.37$ \\
                                 & $512$ & dnf & $0.26 \pm 0.26$ & $\mathbf{0.21 \pm 0.21}$ & oom & $0.40 \pm 0.40$ \\
                                 & $1024$ & dnf & $0.46 \pm 0.46$ & $\mathbf{0.14 \pm 0.14}$ & oom & $0.17 \pm 0.17$ \\
		\bottomrule
	\end{tabular}
	\label{table:result-TimeFID}
\end{table*}

\begin{table*}[t]
	\caption{KL divergence between spatial marginal distributions. TDDPM in unconditional mode except for sequence length $128^*$. Then TDDPM is trained on one quarter of the map and KL is calculated over full dataset including both in- and out-of-distribution data. Lower is better.}
	\centering
	\small
	\begin{tabular}{cl ccccc}
		\addlinespace[0.5em]
		\toprule
		Metric & Len. & TimeGAN & TimeVAE & COSCI-GAN & DiffusionTS & TDDPM \\
		\midrule
\addlinespace[0.5em]
    \multirow{7}{*}{KL(real$\,\vert\vert\,$synthetic)} 
             & $24$ & $2.78$ & $1.51$ & $1.07$ & $\mathbf{0.75}$ & $\mathbf{0.75}$ \\
             & $32$ & $1.41$ & $1.49$ & $1.30$ & $0.70$ & $\mathbf{0.69}$ \\
             & $64$ & $1.40$ & $1.25$ & $1.00$ & $\mathbf{0.68}$ & $0.72$ \\
             & $128$ & $1.77$ & $1.08$ & $1.01$ & $0.80$ & $\mathbf{0.76}$ \\
             & $128^*$ & - & - & - & - & $\mathbf{0.41}$ \\
             & $256$ & $2.73$ & $1.47$ & $1.78$ & $2.23$ & $\mathbf{1.01}$  \\
             & $512$ & dnf & $3.57$ & $2.88$ & dnf & $\mathbf{1.19}$ \\
             & $1024$ & dnf & $6.40$ & $3.98$ & dnf & $\mathbf{0.61}$ \\
             \addlinespace[0.5em]
    \multirow{7}{*}{KL(synthetic$\,\vert\vert\,$real)} 
            & $24$ & $3.31$ & $1.83$ & $1.65$ & $\mathbf{0.94}$ & $\mathbf{0.94}$ \\
            & $32$ & $1.80$ & $1.92$ & $1.82$ & $\mathbf{0.85}$ & $0.86$ \\
            & $64$ & $1.83$ & $1.71$ & $1.44$ & $\mathbf{0.73}$ & $0.89$ \\
             & $128$ & $2.01$ & $1.46$ & $1.49$ & $\mathbf{0.77}$ & $1.02$ \\
             & $128^*$ & - & - & - & - & $\mathbf{0.45}$ \\
             & $256$ & $4.08$ & $2.07$ & $2.93$ & $1.79$ & $\mathbf{1.25}$ \\
            & $512$ & dnf & $6.05$ & $4.98$ & dnf & $\mathbf{1.22}$ \\
            & $1024$ & dnf & $8.76$ & $6.50$ & dnf & $\mathbf{0.61}$ \\
    \bottomrule
	\end{tabular}
	\label{table:KL}
\end{table*}

\subsection{Large-scale human mobility conditional generative model}

In these experiments, we validate the conditional method for generating large spatio-temporal data.
For this we use a quadrant of GeoLife (Large) which is 40 $\text{km}^2$.
Journeys with observations that are longer than 10 seconds apart are split into separate trajectories.
Observation are limited to the geographical area of longitude in the range (116.30, 116.35) and latitude in (39.90, 39.95), any trajectory with an observation outside that area is trimmed to only contain observations in this area.
The minimum window size is 24 and any trajectory with fewer than 24 observations in the geographical area are dropped.
Time between events is assumed constant and altitude is dropped, meaning that that latitude and longitude are observed at each step.
After having trained on the region, we use the model to generate trajectories for the entire training region, see Figure \ref{fig:interpolation-geolife}.
We observe that the synthetic trajectories are similar to the training data and when used to calculate a spatial marginalized distribution, the resulting distribution is close to indistinguishable from the original.

\subsection{Generalizing to new environments}

In this experiment, we generate heatmaps outside of the area used for training the model.
We expand the geographical area by four times, going from a physical area of 40 $\text{km}^2$ to an area of 162 $\text{km}^2$.
Next, we generate heatmaps for all the regions.
In regions outside the geographical area, the heatmaps are generated with data previously not observed by the model.
The model is sampled once for each heatmap, the number of samples drawn proportional to the total number of observations in the area of the heatmap.
To evaluate the quality of the resulting synthetic trajectories, we calculate a global heatmap with all synthetic trajectories from all regions.
This heatmap can then be compared to a heatmap of the original, previously unobserved, data.
The heatmaps and trajectories are all shown in Figure \ref{fig:generalization-geolife}, the upper left figure shows the raw trajectories data, the top right figure shows the chosen regions overlaying a heatmap of the raw trajectories, the resulting synthetic trajectories are shown bottom left and in the bottom right the heatmap of the synthetic trajectories.
Finally, a proof of concept of what-if-scenario modelling is shown in Figure \ref{fig:what-if-scenario}.

\section{Conclusion}
Scaling time-series generative models to long sequences and large-scale settings, not least for mobility data applications, has been very challenging. By conditioning denoising diffusion probabilistic models on a spatial marginal distribution, we demonstrate that our approach TDDPM scales to problem sizes far beyond current state-of-the-art, without compromising fidelity in the generated trajectories. Moreover, TDDPM stays on-par in performance on unconditional tasks at smaller scales while including less induction bias on trajectories and being more efficient at both training time and sampling time than comparable high-fidelity approaches. Finally, high-quality out-of-distribution generalization is demonstrated at scale, including for what-if scenario modeling of road traffic. The accompanying comprehensive benchmark invite continued improvements as future work and a range of applications.


\newpage
\appendix

\section{Appendix / supplemental material}

\subsection{Architecture Additional Details}

Positional encoding~\cite{ho2020denoising}:
\begin{align}
  \text{PE}_{(pos, 2i)} &= \sin \left(-e^i \frac{\log(10000)}{\frac{d}{2} - 1} \right) \\
  \text{PE}_{(pos, 2i + 1)} &= \cos \left(-e^i \frac{\log(10000)}{\frac{d}{2} - 1} \right)
  \label{eq:positional-encoding}
\end{align}

\label{appendix:encoder-input}
The input to the transformer encoder is:
\begin{itemize}
	\item L input tokens, each token corresponding to a time point in the noisy sequence. \textbf{Note:} The token size depends on the number of features of the dataset. It is a concatenation of:
	\begin{itemize}
		\item $x \in \mathbb{R}^{DN}$, each corresponding to a dimension observed at each time point encoded using positional encoding
		\item $x \in \mathbb{R}^N$, the time point encoded using the positional encoding introduced in \cite{vaswani2017attention}
		\item $x \in \mathbb{R}^N$, a learned vector encoding denoting that this is a token that corresponds to a noisy sequence
	\end{itemize}
	\item \textbf{Optional conditional information:} 64 tokens, each corresponding to a patch of the heatmap and being a concatenation of:
	\begin{itemize}
		\item $x \in \mathbb{R}^N$, corresponding to the x position of the patch. Encoded using positional encoding~\cite{vaswani2017attention}.
		\item $x \in \mathbb{R}^N$, corresponding to the y position of the patch. Encoded using positional encoding~\cite{vaswani2017attention}.
		\item $x \in \mathbb{R}^N$, corresponding to the intensity of the heatmap. Encoded using a linear projection~\cite{dosovitskiy2020image}.
		\item $x \in \mathbb{R}^N$, a learned vector encoding denoting that this is a token that corresponds to the conditional information
	\end{itemize}
	\item A token encoding the current denoising step:
	\begin{itemize}
		\item $x \in \mathbb{R}^{N(D+1)}$, the denoising step encoded using positional encoding~\cite{vaswani2017attention}
		\item $x \in \mathbb{R}^N$, a learned vector encoding denoting that this is a token that corresponds to the denoising step
	\end{itemize}
\end{itemize}

\subsection{Additional Results}

\begin{table*}[t]
	\caption{Evaluation of unconditional generation using TSTR. Lower is better. Several experiments ran out of 24GB of VRAM (\emph{oom}) or did not finish (\emph{dnf}) in under 100 hours.}
	\centering
	\small
	\begin{tabular}{ll ccccc}
		\addlinespace[0.5em]
		\toprule
		Dataset & Len. & TimeGAN & TimeVAE & COSCI-GAN & DiffusionTS & Ours \\
		\midrule
		\addlinespace[0.5em]
\multirow{7}{*}{Stock} & $24$ & $\mathbf{0.01 \pm 0.01}$ & $0.03 \pm 0.03$ & $0.02 \pm 0.02$ & $0.02 \pm 0.02$ & $0.02 \pm 0.02$ \\
                       & $32$ & $\mathbf{0.01 \pm 0.01}$ & $\mathbf{0.01 \pm 0.01}$ & $\mathbf{0.01 \pm 0.01}$ & $0.02 \pm 0.02$ & $\mathbf{0.01 \pm 0.01}$ \\
                       & $64$ & $\mathbf{0.01 \pm 0.01}$ & $\mathbf{0.01 \pm 0.01}$ & $0.02 \pm 0.02$ & $\mathbf{0.01 \pm 0.01}$ & $0.02 \pm 0.02$ \\
                       & $128$ & $0.02 \pm 0.02$ & $0.02 \pm 0.02$ & $\mathbf{0.01 \pm 0.01}$ & $\mathbf{0.01 \pm 0.01}$ & $\mathbf{0.01 \pm 0.01}$ \\
                       & $256$ & $0.04 \pm 0.04$ & $0.03 \pm 0.03$ & $\mathbf{0.01 \pm 0.01}$ & $0.02 \pm 0.02$ & $\mathbf{0.01 \pm 0.01}$ \\
                       & $512$ & dnf & $0.02 \pm 0.02$ & $0.02 \pm 0.02$ & oom & $\mathbf{0.01 \pm 0.01}$ \\
                       & $1024$ & dnf & $0.04 \pm 0.04$ & $\mathbf{0.01 \pm 0.01}$ & oom & $\mathbf{0.01 \pm 0.01}$ \\
\addlinespace[0.5em]
\multirow{7}{*}{Energy} & $24$ & $0.07 \pm 0.07$ & $0.06 \pm 0.06$ & $\mathbf{0.05 \pm 0.05}$ & $\mathbf{0.05 \pm 0.05}$ & $\mathbf{0.05 \pm 0.05}$ \\
                        & $32$ & $0.06 \pm 0.06$ & $0.06 \pm 0.06$ & $\mathbf{0.05 \pm 0.05}$ & $\mathbf{0.05 \pm 0.05}$ & $\mathbf{0.05 \pm 0.05}$ \\
                        & $64$ & $0.06 \pm 0.06$ & $0.06 \pm 0.06$ & $\mathbf{0.05 \pm 0.05}$ & $\mathbf{0.05 \pm 0.05}$ & $\mathbf{0.05 \pm 0.05}$ \\
                        & $128$ & $0.09 \pm 0.09$ & $0.06 \pm 0.06$ & $\mathbf{0.05 \pm 0.05}$ & $\mathbf{0.05 \pm 0.05}$ & $\mathbf{0.05 \pm 0.05}$ \\
                        & $256$ & $0.21 \pm 0.21$ & $0.06 \pm 0.06$ & $\mathbf{0.05 \pm 0.05}$ & $\mathbf{0.05 \pm 0.05}$ & $\mathbf{0.05 \pm 0.05}$ \\
                        & $512$ & dnf & $\mathbf{0.05 \pm 0.05}$ & $0.06 \pm 0.06$ & oom & $0.06 \pm 0.06$ \\
                        & $1024$ & dnf & oom & $0.13 \pm 0.13$ & oom & $\mathbf{0.05 \pm 0.05}$ \\
\addlinespace[0.5em]
\multirow{7}{*}{Solar} & $24$ & oom & $\mathbf{0.01 \pm 0.01}$ & $0.92 \pm 0.92$ & $0.07 \pm 0.07$ & $0.07 \pm 0.07$ \\
                       & $32$ & oom & $\mathbf{0.01 \pm 0.01}$ & $0.54 \pm 0.54$ & $0.07 \pm 0.07$ & $0.07 \pm 0.07$ \\
                       & $64$ & oom & $\mathbf{0.01 \pm 0.01}$ & $0.07 \pm 0.07$ & $0.07 \pm 0.07$ & $0.07 \pm 0.07$ \\
                       & $128$ & oom & $\mathbf{0.01 \pm 0.01}$ & $0.80 \pm 0.80$ & $\mathbf{0.01 \pm 0.01}$ & $0.07 \pm 0.07$ \\
                       & $256$ & oom & oom & $0.92 \pm 0.92$ & $\mathbf{0.07 \pm 0.07}$ & $\mathbf{0.07 \pm 0.07}$ \\
                       & $512$ & oom & oom & $0.93 \pm 0.93$ & oom & $\mathbf{0.07 \pm 0.07}$ \\
                       & $1024$ & oom & oom & dnf & oom & $\mathbf{0.07 \pm 0.07}$ \\
\addlinespace[0.5em]
\multirow{7}{*}{Electricity} & $24$ & oom & $\mathbf{0.00 \pm 0.00}$ & dnf & $\mathbf{0.00 \pm 0.00}$ & $\mathbf{0.00 \pm 0.00}$ \\
                             & $32$ & oom & $\mathbf{0.00 \pm 0.00}$ & dnf & $\mathbf{0.00 \pm 0.00}$ & $\mathbf{0.00 \pm 0.00}$ \\
                             & $64$ & oom & $\mathbf{0.00 \pm 0.00}$ & dnf & $\mathbf{0.00 \pm 0.00}$ & $\mathbf{0.00 \pm 0.00}$ \\
                             & $128$ & oom & oom & dnf & $0.01 \pm 0.01$ & $\mathbf{0.00 \pm 0.00}$ \\
                             & $256$ & oom & oom & dnf & $\mathbf{0.00 \pm 0.00}$ & $\mathbf{0.00 \pm 0.00}$ \\
                             & $512$ & oom & oom & dnf & oom & $\mathbf{0.00 \pm 0.00}$ \\
                             & $1024$ & oom & oom & dnf & oom & $\mathbf{0.00 \pm 0.00}$ \\
\addlinespace[0.5em]
\multirow{7}{*}{ATC} & $24$ & dnf & $0.31 \pm 0.31$ & dnf & $\mathbf{0.04 \pm 0.04}$ & $0.09 \pm 0.09$ \\
                     & $32$ & dnf & $0.27 \pm 0.27$ & dnf & $\mathbf{0.04 \pm 0.04}$ & $\mathbf{0.04 \pm 0.04}$ \\
                     & $64$ & dnf & $0.26 \pm 0.26$ & dnf & $\mathbf{0.04 \pm 0.04}$ & $\mathbf{0.04 \pm 0.04}$ \\
                     & $128$ & dnf & oom & dnf & $\mathbf{0.04 \pm 0.04}$ & $\mathbf{0.04 \pm 0.04}$ \\
                     & $256$ & dnf & oom & dnf & $\mathbf{0.04 \pm 0.04}$ & $\mathbf{0.04 \pm 0.04}$ \\
                     & $512$ & dnf & oom & dnf & oom & $\mathbf{0.04 \pm 0.04}$ \\
                     & $1024$ & dnf & oom & dnf & oom & $\mathbf{0.05 \pm 0.05}$ \\
\addlinespace[0.5em]
\multirow{7}{*}{Geolife (Small)} & $24$ & $0.12 \pm 0.12$ & $0.10 \pm 0.10$ & $0.11 \pm 0.11$ & $0.13 \pm 0.13$ & $\mathbf{0.09 \pm 0.09}$ \\
                                 & $32$ & $\mathbf{0.09 \pm 0.09}$ & $\mathbf{0.09 \pm 0.09}$ & $0.10 \pm 0.10$ & $\mathbf{0.09 \pm 0.09}$ & $\mathbf{0.09 \pm 0.09}$ \\
                                 & $64$ & $0.13 \pm 0.13$ & $\mathbf{0.08 \pm 0.08}$ & $0.11 \pm 0.11$ & $0.09 \pm 0.09$ & $0.10 \pm 0.10$ \\
                                 & $128$ & $\mathbf{0.08 \pm 0.08}$ & $0.12 \pm 0.12$ & $\mathbf{0.08 \pm 0.08}$ & $\mathbf{0.08 \pm 0.08}$ & $\mathbf{0.08 \pm 0.08}$ \\
                                 & $256$ & $0.07 \pm 0.07$ & $\mathbf{0.06 \pm 0.06}$ & $0.07 \pm 0.07$ & $0.07 \pm 0.07$ & $0.07 \pm 0.07$ \\
                                 & $512$ & dnf & $0.05 \pm 0.05$ & $\mathbf{0.04 \pm 0.04}$ & oom & $\mathbf{0.04 \pm 0.04}$ \\
                                 & $1024$ & dnf & $0.10 \pm 0.10$ & $\mathbf{0.02 \pm 0.02}$ & oom & $\mathbf{0.02 \pm 0.02}$ \\

		\bottomrule
	\end{tabular}
	\label{table:result-TSTR}
\end{table*}

\begin{table*}[t]
	\caption{Divergence between spatial marginal distributions. TDDPM in unconditional mode except for sequence length $128^*$. Then TDDPM is trained on one quarter of the map and KL is calculated over full dataset including both in- and out-of-distribution data. Lower is better.}
	\centering
	\small
	\begin{tabular}{cl ccccc}
		\addlinespace[0.5em]
		\toprule
		Dataset & Len. & TimeGAN & TimeVAE & COSCI-GAN & DiffusionTS & TDDPM \\
		\midrule
\addlinespace[0.5em]
    \multirow{7}{*}{Symmetric KL} 
    & $24$ & $3.04$ & $1.67$ & $1.36$ & $\mathbf{0.84}$ & $0.85$ \\
    & $32$ & $1.60$ & $1.70$ & $1.56$ & $\mathbf{0.78}$ & $\mathbf{0.78}$ \\
    & $64$ & $1.61$ & $1.48$ & $1.22$ & $\mathbf{0.71}$ & $0.80$ \\
    & $128$ & $1.89$ & $1.27$ & $1.25$ & $\mathbf{0.79}$ & $0.89$ \\
    & $128^*$ & - & - & - & - & $\mathbf{0.43}$ \\
    & $256$ & $3.40$ & $1.77$ & $2.36$ & $2.01$ & $\mathbf{1.13}$ \\
    & $512$ & dnf & $4.81$ & $3.93$ & oom & $\mathbf{1.20}$ \\
    & $1024$ & dnf & $7.58$ & $5.24$ & oom & $\mathbf{0.61}$ \\
     \addlinespace[0.5em]
    \multirow{7}{*}{JS} 
     & $24$ & $0.44$ & $0.28$ & $0.23$ & $\mathbf{0.16}$ & $\mathbf{0.16}$ \\
     & $32$ & $0.26$ & $0.28$ & $0.26$ & $0.15$ & $\mathbf{0.14}$ \\
     & $64$ & $0.26$ & $0.24$ & $0.20$ & $\mathbf{0.14}$ & $0.15$ \\
     & $128$ & $0.29$ & $0.21$ & $0.21$ & $\mathbf{0.15}$ & $0.16$ \\
     & $128^*$ & - & - & - & - & $\mathbf{0.09}$ \\
     & $256$ & $0.42$ & $0.27$ & $0.34$ & $0.31$ & $\mathbf{0.20}$ \\
     & $512$ & dnf & $0.51$ & $0.48$ & dnf & $\mathbf{0.19}$ \\
     & $1024$ & dnf & $0.66$ & $0.58$ & dnf & $\mathbf{0.11}$ \\
    \bottomrule
	\end{tabular}
	\label{table:divergence}
\end{table*}

\begin{figure}
	\centering
	\includegraphics[width=1\linewidth]{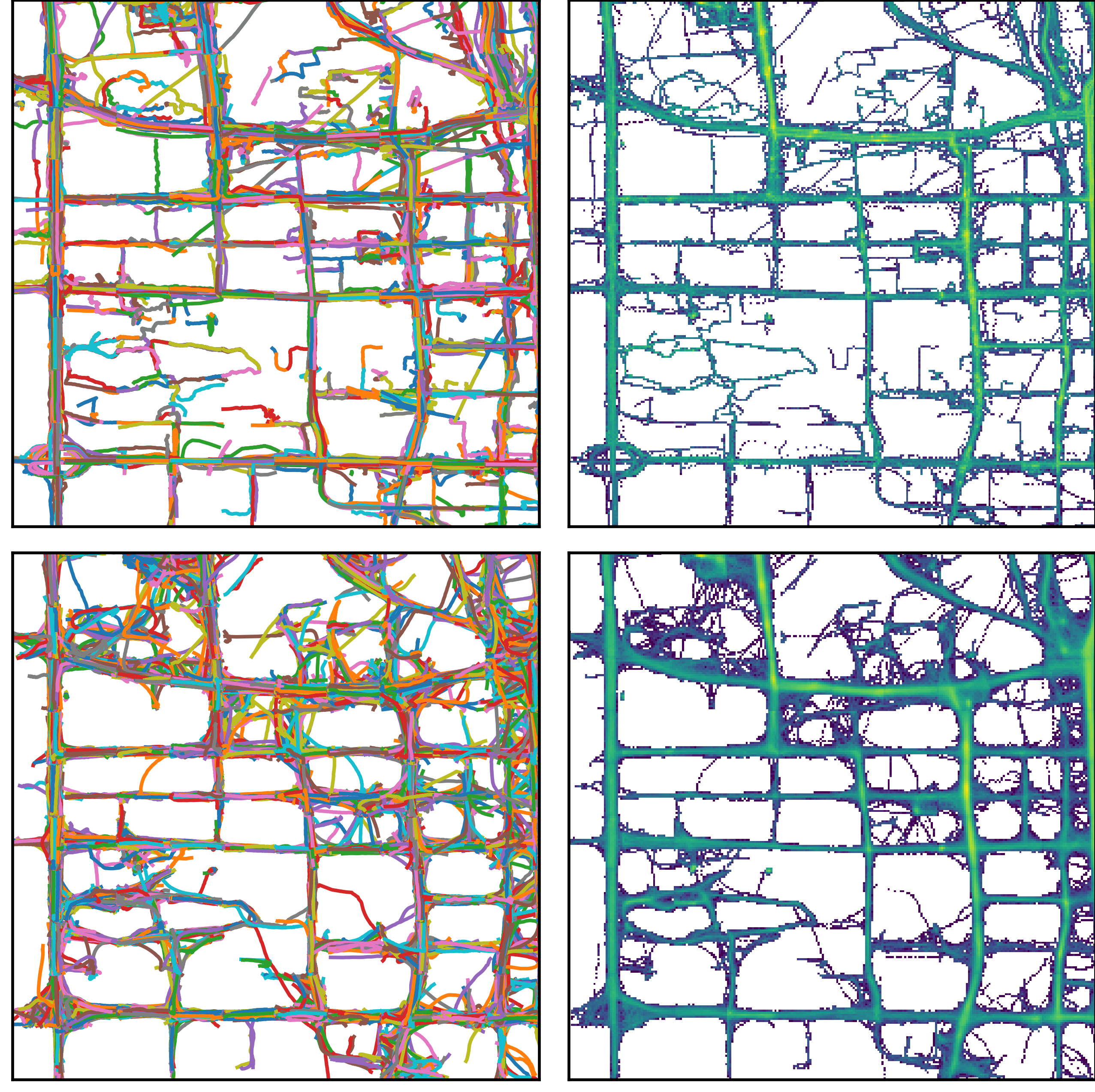}
	\caption{Interpolation experiment. The model has trained on this region is tasked to reconstruct it from the heatmaps. \textit{Top left:} Data from GeoLife used for training. \textit{Top right:} heatmap of training data and areas used for creating query heatmaps for sampling the model,
		\textit{Bottom left:} synthetic trajectories.
		\textit{Bottom right:} heatmap of the synthetic data.}
	\label{fig:interpolation-geolife}
\end{figure}

\begin{figure}
	\centering
	\includegraphics[width=1\linewidth]{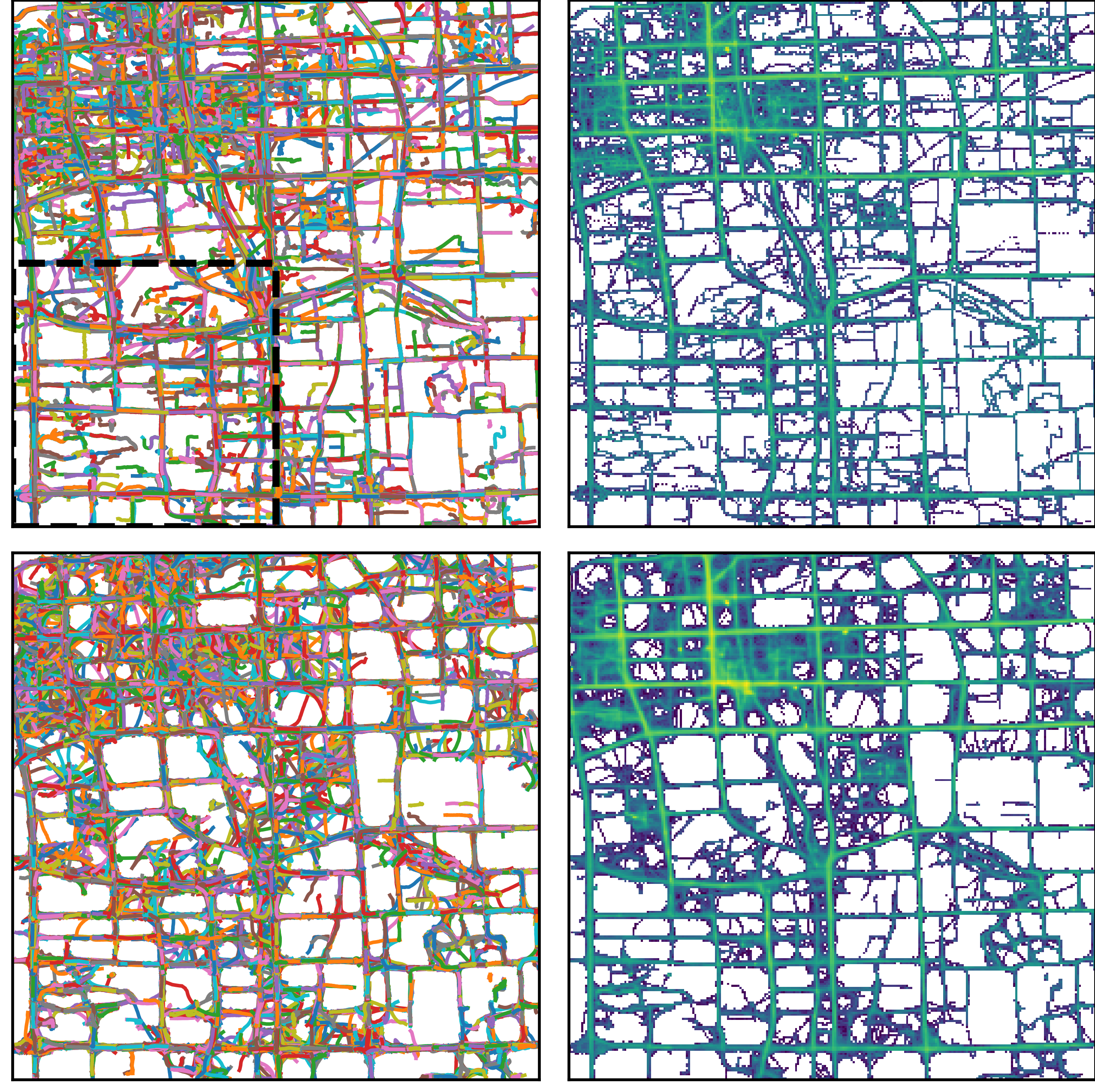}
	\caption{Generalization experiment. The model is trained on the lower left quadrant and used to generate data on the remaining geographical area. \textit{Top left:} Data from GeoLife, the lower quandrant of which used for training. \textit{Top right:} heatmap of training data,
		\textit{Bottom left:} synthetic trajectories.
		\textit{Bottom right:} heatmap of the synthetic data.}
	\label{fig:generalization-geolife}
\end{figure}

\begin{figure}
	\centering
	\includegraphics[width=.95\linewidth]{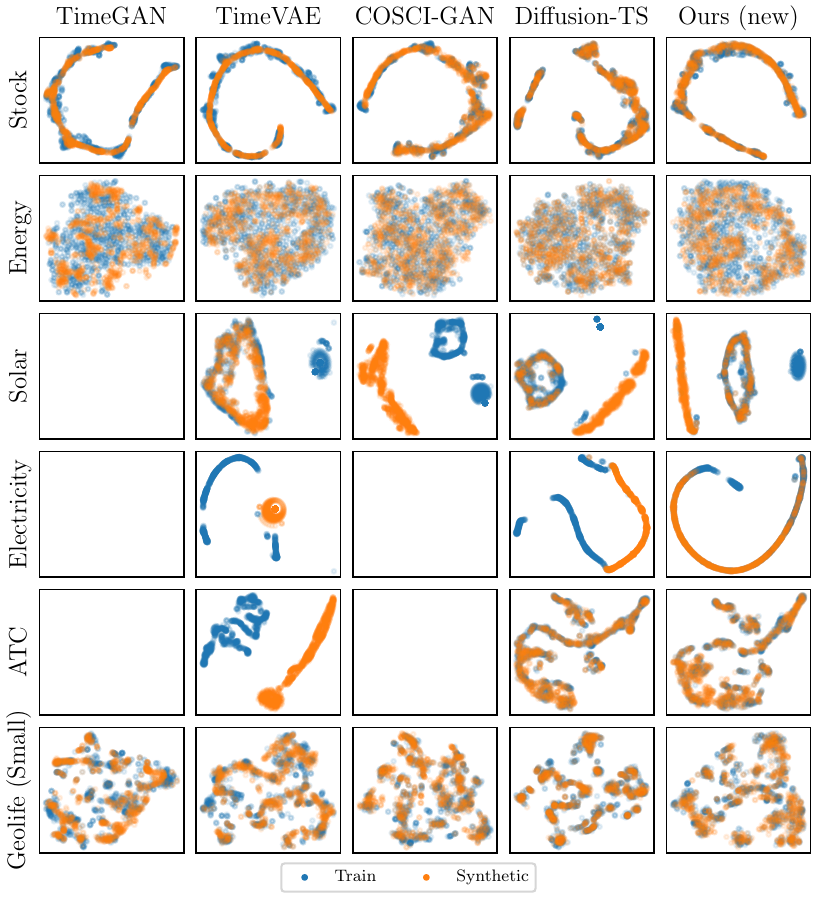}
	\caption{t-SNE analysis for sequence length 64. Blank figures are unsuccessful experiments (Table \ref{table:result-TimeFID}).}
	\label{fig:unconditional-tsne}
\end{figure}

\begin{figure}
	\centering
	\includegraphics[width=.95\linewidth]{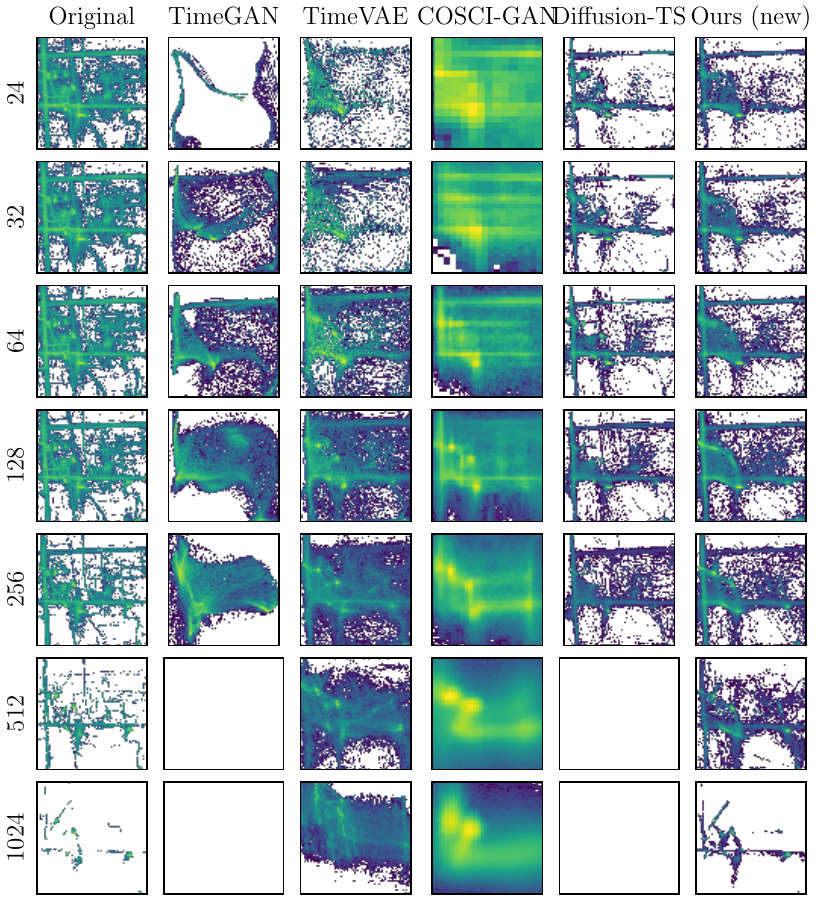}
	\caption{Heatmaps of the unconditionally generated trajectories for GeoLife. Training dataset is shown in the left column. Blank figures are unsuccessful experiments (Table \ref{table:result-TimeFID}).}
	\label{fig:unconditional-geolife}
\end{figure}

\begin{figure}
	\centering
	\includegraphics[width=.95\linewidth]{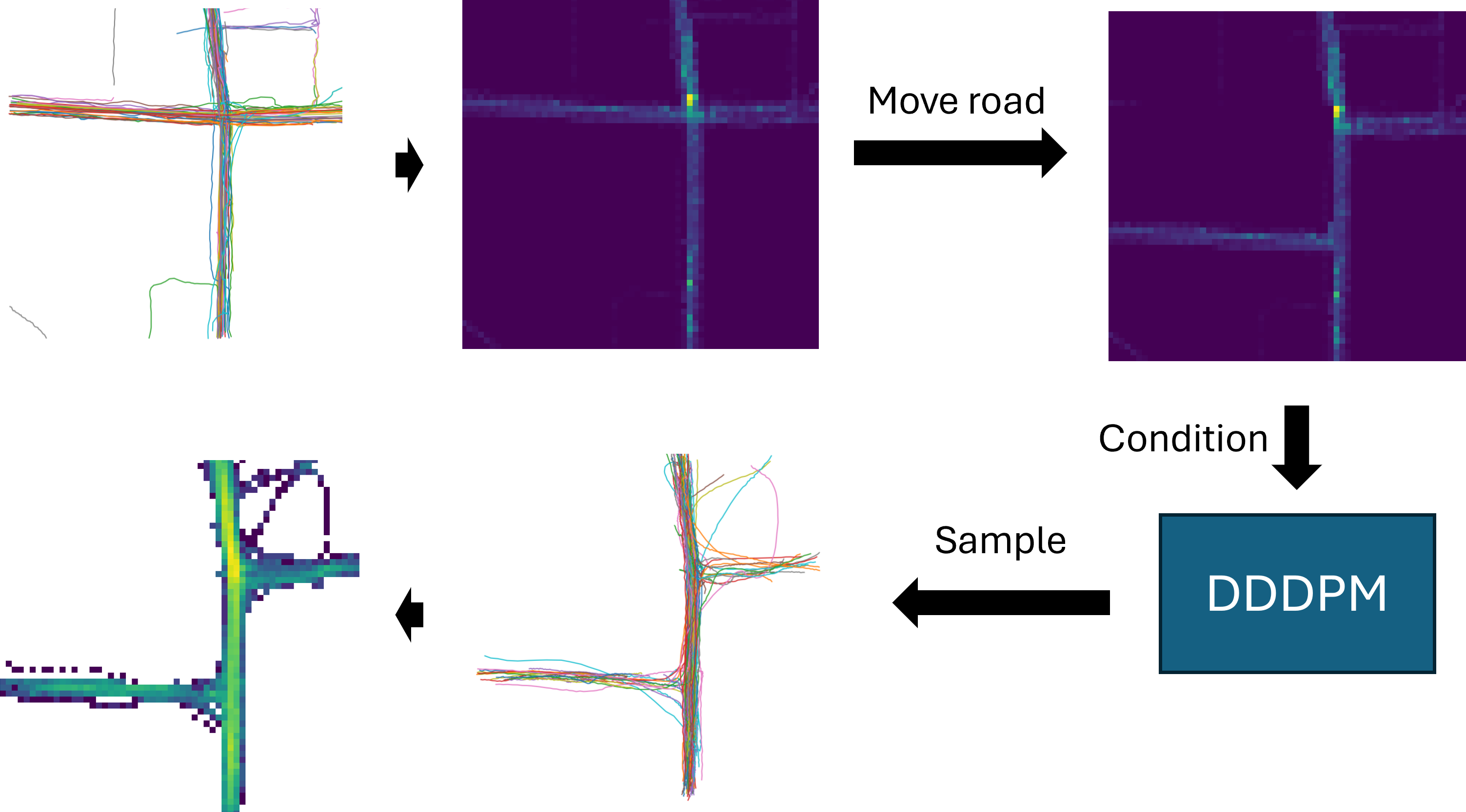}
	\caption{What-if-scenario where a road is removed and added.}
	\label{fig:what-if-scenario}
\end{figure}

\begin{table*}[t]
  \caption{Selected properties of the datasets used in evaluation of unconditional generation. A dataset consists of one or more sequences, each sequence consist of one or more observation across time and at each time point one or more dimensions are observed.}
  \label{table:datasets}
  \centering
  \begin{tabular}{lrrrr}
    \addlinespace[0.5em]
    \toprule
    Name & Number of sequences & Sequence length & Observations & Dimensions \\
    \midrule
    Stock & 1 & 3,685 & 3,865 & 6 \\
    Energy & 1 & 19,735 & 19,735 & 28 \\
    Solar & 137 & 105,121 & 14,401,440 & 1 \\
    Electricity & 370 & [16,032, 140,256] & 41,855,506 & 1 \\
    ATC & 49,688 & [24, 187766] & 47,290,292 & 2 \\
    Geolife (Smaller) & 16,708 & [24, 3,322] & 1,957,039 & 2 \\
    \bottomrule
    \end{tabular}
\end{table*}


\begin{thebibliography}{38}
\providecommand{\natexlab}[1]{#1}
\providecommand{\url}[1]{\texttt{#1}}
\expandafter\ifx\csname urlstyle\endcsname\relax
  \providecommand{\doi}[1]{doi: #1}\else
  \providecommand{\doi}{doi: \begingroup \urlstyle{rm}\Url}\fi

\bibitem[Ilin et~al.(2021)Ilin, Annan-Phan, Tai, Mehra, Hsiang, and
  Blumenstock]{ilin2021public}
Cornelia Ilin, S{\'e}bastien Annan-Phan, Xiao~Hui Tai, Shikhar Mehra, Solomon
  Hsiang, and Joshua~E Blumenstock.
\newblock Public mobility data enables covid-19 forecasting and management at
  local and global scales.
\newblock \emph{Scientific reports}, 11\penalty0 (1):\penalty0 13531, 2021.

\bibitem[Wang et~al.(2022)Wang, Currim, and Ram]{wang2022deep}
Yun Wang, Faiz Currim, and Sudha Ram.
\newblock Deep learning of spatiotemporal patterns for urban mobility
  prediction using big data.
\newblock \emph{Information Systems Research}, 33\penalty0 (2):\penalty0
  579--598, 2022.

\bibitem[Xiong et~al.(2024)Xiong, Li, Zhao, Zhang, Miao, Lv, and
  Wang]{xiong2024trajsgan}
Gang Xiong, Zhishuai Li, Meihua Zhao, Yu~Zhang, Qinghai Miao, Yisheng Lv, and
  Fei-Yue Wang.
\newblock Trajsgan: A semantic-guiding adversarial network for rban trajectory
  generation.
\newblock \emph{IEEE Transactions on Computational Social Systems}, 11\penalty0
  (2):\penalty0 1733--1743, 2024.
\newblock \doi{10.1109/TCSS.2023.3235923}.

\bibitem[Tai et~al.(2022)Tai, Mehra, and Blumenstock]{tai2022mobile}
Xiao~Hui Tai, Shikhar Mehra, and Joshua~E Blumenstock.
\newblock Mobile phone data reveal the effects of violence on internal
  displacement in afghanistan.
\newblock \emph{Nature human behaviour}, 6\penalty0 (5):\penalty0 624--634,
  2022.

\bibitem[Niva et~al.(2023)Niva, Horton, Virkki, Heino, Kosonen, Kallio,
  Kinnunen, Abel, Muttarak, Taka, et~al.]{niva2023world}
Venla Niva, Alexander Horton, Vili Virkki, Matias Heino, Maria Kosonen, Marko
  Kallio, Pekka Kinnunen, Guy~J Abel, Raya Muttarak, Maija Taka, et~al.
\newblock World’s human migration patterns in 2000--2019 unveiled by
  high-resolution data.
\newblock \emph{Nature Human Behaviour}, 7\penalty0 (11):\penalty0 2023--2037,
  2023.

\bibitem[Alessandrini et~al.(2020)Alessandrini, Ghio, Migali,
  et~al.]{alessandrini2020estimating}
Alfredo Alessandrini, Daniela Ghio, Silvia Migali, et~al.
\newblock \emph{Estimating net migration at high spatial resolution}.
\newblock Publications Office of the European Union, 2020.

\bibitem[Ansari et~al.(2024)Ansari, Stella, Turkmen, Zhang, Mercado, Shen,
  Shchur, Rangapuram, Arango, Kapoor, et~al.]{ansari2024chronos}
Abdul~Fatir Ansari, Lorenzo Stella, Caner Turkmen, Xiyuan Zhang, Pedro Mercado,
  Huibin Shen, Oleksandr Shchur, Syama~Sundar Rangapuram, Sebastian~Pineda
  Arango, Shubham Kapoor, et~al.
\newblock Chronos: Learning the language of time series.
\newblock \emph{arXiv preprint arXiv:2403.07815}, 2024.

\bibitem[Della~Valle et~al.(2009)Della~Valle, Ceri, Van~Harmelen, and
  Fensel]{della2009s}
Emanuele Della~Valle, Stefano Ceri, Frank Van~Harmelen, and Dieter Fensel.
\newblock It's a streaming world! reasoning upon rapidly changing information.
\newblock \emph{IEEE Intelligent Systems}, 24\penalty0 (6):\penalty0 83--89,
  2009.

\bibitem[Lana et~al.(2018)Lana, Del~Ser, Velez, and Vlahogianni]{lana2018road}
Ibai Lana, Javier Del~Ser, Manuel Velez, and Eleni~I Vlahogianni.
\newblock Road traffic forecasting: Recent advances and new challenges.
\newblock \emph{IEEE Intelligent Transportation Systems Magazine}, 10\penalty0
  (2):\penalty0 93--109, 2018.

\bibitem[Yoon et~al.(2019{\natexlab{a}})Yoon, Jordon, and van~der
  Schaar]{yoon2018pategan}
Jinsung Yoon, James Jordon, and Mihaela van~der Schaar.
\newblock {PATE}-{GAN}: Generating synthetic data with differential privacy
  guarantees.
\newblock In \emph{International Conference on Learning Representations},
  2019{\natexlab{a}}.
\newblock URL \url{https://openreview.net/forum?id=S1zk9iRqF7}.

\bibitem[Wang et~al.(2023)Wang, Gao, Wu, Jin, Yao, and Li]{wang2023pategail}
Huandong Wang, Changzheng Gao, Yuchen Wu, Depeng Jin, Lina Yao, and Yong Li.
\newblock Pategail: a privacy-preserving mobility trajectory generator with
  imitation learning.
\newblock In \emph{Proceedings of the AAAI Conference on Artificial
  Intelligence}, volume~37, pages 14539--14547, 2023.

\bibitem[Alcaraz and Strodthoff(2023)]{alcaraz2023diffusionbased}
Juan~Lopez Alcaraz and Nils Strodthoff.
\newblock Diffusion-based time series imputation and forecasting with
  structured state space models.
\newblock \emph{Transactions on Machine Learning Research}, 2023.
\newblock ISSN 2835-8856.
\newblock URL \url{https://openreview.net/forum?id=hHiIbk7ApW}.

\bibitem[Yoon et~al.(2019{\natexlab{b}})Yoon, Jarrett, and Van~der
  Schaar]{yoon2019time}
Jinsung Yoon, Daniel Jarrett, and Mihaela Van~der Schaar.
\newblock {Time-series Generative Adversarial Networks}.
\newblock \emph{Advances in neural information processing systems (NeurIPS)},
  32, 2019{\natexlab{b}}.

\bibitem[Seyfi et~al.(2022)Seyfi, Rajotte, and Ng]{seyfi2022generating}
Ali Seyfi, Jean-Francois Rajotte, and Raymond Ng.
\newblock Generating multivariate time series with {CO}mmon {S}ource
  {C}oord{I}nated {GAN} ({COSCI-GAN}).
\newblock \emph{Advances in Neural Information Processing Systems},
  35:\penalty0 32777--32788, 2022.

\bibitem[Yuan and Qiao(2024)]{yuan2024diffusionts}
Xinyu Yuan and Yan Qiao.
\newblock Diffusion-{TS}: Interpretable diffusion for general time series
  generation.
\newblock In \emph{The Twelfth International Conference on Learning
  Representations}, 2024.
\newblock URL \url{https://openreview.net/forum?id=4h1apFjO99}.

\bibitem[Alaa et~al.(2022)Alaa, Van~Breugel, Saveliev, and van~der
  Schaar]{alaa2022faithful}
Ahmed Alaa, Boris Van~Breugel, Evgeny~S. Saveliev, and Mihaela van~der Schaar.
\newblock How faithful is your synthetic data? {S}ample-level metrics for
  evaluating and auditing generative models.
\newblock In Kamalika Chaudhuri, Stefanie Jegelka, Le~Song, Csaba Szepesvari,
  Gang Niu, and Sivan Sabato, editors, \emph{Proceedings of the 39th
  International Conference on Machine Learning}, volume 162 of
  \emph{Proceedings of Machine Learning Research}, pages 290--306. PMLR, 17--23
  Jul 2022.

\bibitem[Wu et~al.(2021)Wu, Gao, and Zha]{wu2021bridging}
Qitian Wu, Rui Gao, and Hongyuan Zha.
\newblock Bridging explicit and implicit deep generative models via neural
  stein estimators.
\newblock \emph{Advances in Neural Information Processing Systems},
  34:\penalty0 11274--11286, 2021.

\bibitem[Esteban et~al.(2017)Esteban, Hyland, and R{\"a}tsch]{esteban2017real}
Crist{\'o}bal Esteban, Stephanie~L Hyland, and Gunnar R{\"a}tsch.
\newblock {Real-valued (Medical) Time Series Generation with Recurrent
  Conditional GANs}.
\newblock \emph{arXiv preprint arXiv:1706.02633}, 2017.

\bibitem[Jeon et~al.(2022)Jeon, Kim, Song, Cho, and Park]{jeon2022gt}
Jinsung Jeon, Jeonghak Kim, Haryong Song, Seunghyeon Cho, and Noseong Park.
\newblock Gt-gan: General purpose time series synthesis with generative
  adversarial networks.
\newblock \emph{Advances in Neural Information Processing Systems},
  35:\penalty0 36999--37010, 2022.

\bibitem[Tashiro et~al.(2021)Tashiro, Song, Song, and Ermon]{tashiro2021csdi}
Yusuke Tashiro, Jiaming Song, Yang Song, and Stefano Ermon.
\newblock Csdi: Conditional score-based diffusion models for probabilistic time
  series imputation.
\newblock In M.~Ranzato, A.~Beygelzimer, Y.~Dauphin, P.S. Liang, and J.~Wortman
  Vaughan, editors, \emph{Advances in Neural Information Processing Systems},
  volume~34, pages 24804--24816. Curran Associates, Inc., 2021.
\newblock URL
  \url{https://proceedings.neurips.cc/paper_files/paper/2021/file/cfe8504bda37b575c70ee1a8276f3486-Paper.pdf}.

\bibitem[Shen and Kwok(2023)]{shen2023non}
Lifeng Shen and James Kwok.
\newblock Non-autoregressive conditional diffusion models for time series
  prediction.
\newblock In \emph{International Conference on Machine Learning}, pages
  31016--31029. PMLR, 2023.

\bibitem[Dai et~al.(2023)Dai, Yang, Liu, Liu, and Liu]{dai2023timeddpm}
Yun Dai, Chao Yang, Kaixin Liu, Angpeng Liu, and Yi~Liu.
\newblock Timeddpm: Time series augmentation strategy for industrial soft
  sensing.
\newblock \emph{IEEE Sensors Journal}, 2023.

\bibitem[Feng et~al.(2024)Feng, Miao, Zhang, and Zhao]{feng2024latent}
Shibo Feng, Chunyan Miao, Zhong Zhang, and Peilin Zhao.
\newblock Latent diffusion transformer for probabilistic time series
  forecasting.
\newblock In \emph{Proceedings of the AAAI Conference on Artificial
  Intelligence}, volume~38, pages 11979--11987, 2024.

\bibitem[Desai et~al.(2021)Desai, Freeman, Beaver, and Wang]{desai2021timevae}
Abhyuday Desai, Cynthia Freeman, Ian Beaver, and Zuhui Wang.
\newblock {TimeVAE}: A variational auto-encoder for multivariate time series
  generation.
\newblock \emph{arXiv preprint arXiv:2111.08095}, 2021.

\bibitem[Vaswani et~al.(2017)Vaswani, Shazeer, Parmar, Uszkoreit, Jones, Gomez,
  Kaiser, and Polosukhin]{vaswani2017attention}
Ashish Vaswani, Noam Shazeer, Niki Parmar, Jakob Uszkoreit, Llion Jones,
  Aidan~N Gomez, \L~ukasz Kaiser, and Illia Polosukhin.
\newblock {Attention is All you Need}.
\newblock In I.~Guyon, U.~Von Luxburg, S.~Bengio, H.~Wallach, R.~Fergus,
  S.~Vishwanathan, and R.~Garnett, editors, \emph{Advances in Neural
  Information Processing Systems}, volume~30, 2017.

\bibitem[Dosovitskiy et~al.(2020)Dosovitskiy, Beyer, Kolesnikov, Weissenborn,
  Zhai, Unterthiner, Dehghani, Minderer, Heigold, Gelly,
  et~al.]{dosovitskiy2020image}
Alexey Dosovitskiy, Lucas Beyer, Alexander Kolesnikov, Dirk Weissenborn,
  Xiaohua Zhai, Thomas Unterthiner, Mostafa Dehghani, Matthias Minderer, Georg
  Heigold, Sylvain Gelly, et~al.
\newblock An image is worth 16x16 words: Transformers for image recognition at
  scale.
\newblock \emph{arXiv preprint arXiv:2010.11929}, 2020.

\bibitem[Feng et~al.(2020)Feng, Yang, Xu, Yu, Wang, and Li]{feng2020learning}
Jie Feng, Zeyu Yang, Fengli Xu, Haisu Yu, Mudan Wang, and Yong Li.
\newblock Learning to simulate human mobility.
\newblock In \emph{Proceedings of the 26th ACM SIGKDD International Conference
  on Knowledge Discovery \& Data Mining}, KDD '20, page 3426–3433, New York,
  NY, USA, 2020. Association for Computing Machinery.
\newblock ISBN 9781450379984.
\newblock \doi{10.1145/3394486.3412862}.
\newblock URL \url{https://doi.org/10.1145/3394486.3412862}.

\bibitem[Jiang et~al.(2023)Jiang, Zhao, Wang, and Jiang]{jiang2023continuous}
Wenjun Jiang, Wayne~Xin Zhao, Jingyuan Wang, and Jiawei Jiang.
\newblock Continuous trajectory generation based on two-stage gan.
\newblock In \emph{{AAAI}}. {AAAI} Press, 2023.

\bibitem[Han(2024)]{han2024enhanced}
Lingyun Han.
\newblock Enhanced generation of human mobility trajectory with multiscale
  model.
\newblock In Biao Luo, Long Cheng, Zheng-Guang Wu, Hongyi Li, and Chaojie Li,
  editors, \emph{Neural Information Processing}, pages 309--323, Singapore,
  2024. Springer Nature Singapore.
\newblock ISBN 978-981-99-8178-6.

\bibitem[Wang et~al.(2024)Wang, Zheng, Liang, Liu, and Song]{wang2024cola}
Yu~Wang, Tongya Zheng, Yuxuan Liang, Shunyu Liu, and Mingli Song.
\newblock Cola: Cross-city mobility transformer for human trajectory
  simulation.
\newblock In \emph{Proceedings of the ACM on Web Conference 2024}, pages
  3509--3520, 2024.

\bibitem[Ho et~al.(2020)Ho, Jain, and Abbeel]{ho2020denoising}
Jonathan Ho, Ajay Jain, and Pieter Abbeel.
\newblock {Denoising Diffusion Probabilistic Models}.
\newblock In \emph{{Advances in Neural Information Processing Systems}},
  volume~33, pages 6840--6851. Curran Associates, Inc., 2020.

\bibitem[Xiao et~al.(2022)Xiao, Kreis, and Vahdat]{xiao2021tackling}
Zhisheng Xiao, Karsten Kreis, and Arash Vahdat.
\newblock Tackling the generative learning trilemma with denoising diffusion
  gans.
\newblock 2022.

\bibitem[Jeha et~al.(2021)Jeha, Bohlke-Schneider, Mercado, Kapoor, Nirwan,
  Flunkert, Gasthaus, and Januschowski]{jeha2021psa}
Paul Jeha, Michael Bohlke-Schneider, Pedro Mercado, Shubham Kapoor,
  Rajbir~Singh Nirwan, Valentin Flunkert, Jan Gasthaus, and Tim Januschowski.
\newblock {PSA-GAN: Progressive Self Attention GANs for Synthetic Time Series}.
\newblock In \emph{International Conference on Learning Representations
  (ICLR)}, 2021.

\bibitem[Br{\v{s}}{\v{c}}i{\'c} et~al.(2013)Br{\v{s}}{\v{c}}i{\'c}, Kanda,
  Ikeda, and Miyashita]{brvsvcic2013person}
Dra{\v{z}}en Br{\v{s}}{\v{c}}i{\'c}, Takayuki Kanda, Tetsushi Ikeda, and
  Takahiro Miyashita.
\newblock Person tracking in large public spaces using 3-d range sensors.
\newblock \emph{IEEE Transactions on Human-Machine Systems}, 43\penalty0
  (6):\penalty0 522--534, 2013.

\bibitem[Zheng et~al.(2011)Zheng, Fu, Xie, Ma, and Li]{zheng2011geolife}
Yu~Zheng, Hao Fu, Xing Xie, Wei-Ying Ma, and Quannan Li.
\newblock \emph{Geolife GPS trajectory dataset - User Guide}, geolife gps
  trajectories 1.1 edition, July 2011.
\newblock URL
  \url{https://www.microsoft.com/en-us/research/publication/geolife-gps-trajectory-dataset-user-guide/}.

\bibitem[Heusel et~al.(2017)Heusel, Ramsauer, Unterthiner, Nessler, and
  Hochreiter]{heusel2017gans}
Martin Heusel, Hubert Ramsauer, Thomas Unterthiner, Bernhard Nessler, and Sepp
  Hochreiter.
\newblock Gans trained by a two time-scale update rule converge to a local nash
  equilibrium.
\newblock In I.~Guyon, U.~Von Luxburg, S.~Bengio, H.~Wallach, R.~Fergus,
  S.~Vishwanathan, and R.~Garnett, editors, \emph{Advances in Neural
  Information Processing Systems}, volume~30. Curran Associates, Inc., 2017.

\bibitem[Franceschi et~al.(2019)Franceschi, Dieuleveut, and
  Jaggi]{franceschi2019unsupervised}
Jean-Yves Franceschi, Aymeric Dieuleveut, and Martin Jaggi.
\newblock Unsupervised scalable representation learning for multivariate time
  series.
\newblock In H.~Wallach, H.~Larochelle, A.~Beygelzimer, F.~d\textquotesingle
  Alch\'{e}-Buc, E.~Fox, and R.~Garnett, editors, \emph{Advances in Neural
  Information Processing Systems}, volume~32. Curran Associates, Inc., 2019.

\bibitem[Higham(1988)]{higham1988computing}
Nicholas~J Higham.
\newblock Computing a nearest symmetric positive semidefinite matrix.
\newblock \emph{Linear algebra and its applications}, 103:\penalty0 103--118,
  1988.

\end{thebibliography}
\end{document}